%% file: main.tex
\documentclass{article}

 \usepackage[preprint]{neurips_2026}

\usepackage[utf8]{inputenc} 
\usepackage[T1]{fontenc}    
\usepackage{hyperref}       
\usepackage{url}            
\usepackage{booktabs}       
\usepackage{amsfonts}       
\usepackage{nicefrac}       
\usepackage{microtype}      
\usepackage{xcolor}         

\usepackage{amsmath}
\usepackage{subcaption}
\usepackage{graphicx}
\usepackage{multirow}
\usepackage{algorithm}
\usepackage{algpseudocode}

\usepackage{enumitem}

\setlist[itemize]{leftmargin=*}
\setlist[enumerate]{leftmargin=*}

\title{
Bringing Value Models Back: Generative Critics for Value Modeling in LLM Reinforcement Learning
}

\author{
  Zikang Shan$^{1,2,}$
  \thanks{Work done at Microsoft Research Asia; email to \texttt{<shanzikang@stu.pku.edu.cn>}; $^\dagger$correspondence to Li Zhao \texttt{<lizo@microsoft.com>}, Han Zhong \texttt{<han.zhong@sjtu.edu.cn>} and Liwei Wang \texttt{<wanglw@cis.pku.edu.cn>}.} \quad
  Han Zhong$^{3\,\dagger}$ \quad
  Liwei Wang$^{1\,\dagger}$ \quad
  Li Zhao$^{2\,\dagger}$
  \\[6pt]
  \small
  $^{1}$Peking University \quad
  $^{2}$Microsoft Research Asia \quad
  $^{3}$Shanghai Jiao Tong University
}

\begin{document}

\maketitle

\input{sections/abstract}

\section{Introduction}
\input{sections/introduction}

\section{Related Works}
\input{sections/related_works}

\section{Preliminaries}
\input{sections/preliminaries}

\section{Limitations of Discriminative Critics}
\label{sec:limitations}
\input{sections/limitations}

\section{Generative Actor-Critic}
\label{sec:method}
\input{sections/method}

\section{Experiments}
\label{sec:exp}
\input{sections/experiments}

\section{Limitations and Future Work}
\label{sec:conclusion}
\input{sections/conclusion}

\bibliographystyle{unsrt}
\bibliography{ref}

\newpage
\appendix

\section{Background on Representation Complexity}
\label{app:complexity_background}
\input{appendices/theory}

\section{Implementation Details}
\label{app:implementation}
\input{appendices/implementation}

\section{Ablation of the SFT Stage}
\label{app:sft_ablation}
\input{appendices/sft_ablation}

\section{Multi-Seed Robustness}
\label{app:robustness}
\input{appendices/robustness}

\section{Computational Cost Analysis of GenAC}
\label{app:cost}
\input{appendices/cost}

\section{Broader Impact}
\label{app:impact}
\input{appendices/broader_impact}

\end{document}

%% file: sections/abstract.tex
\begin{abstract}
Credit assignment is a central challenge in reinforcement learning (RL). Classical actor-critic methods address this challenge through fine-grained advantage estimation based on a learned value function. However, learned value models are often avoided in modern large language model (LLM) RL because conventional discriminative critics are difficult to train reliably. We revisit value modeling and argue that this difficulty is partly due to limited expressiveness. In particular, representation complexity theory suggests that value functions can be hard to approximate under the one-shot prediction paradigm used by existing value models, and our scaling experiments show that such critics do not improve reliably with scale. Motivated by this observation, we propose Generative Actor-Critic (GenAC), which replaces one-shot scalar value prediction with a generative critic that performs chain-of-thought reasoning before producing a value estimate. We further introduce In-Context Conditioning, which helps the critic remain calibrated to the current actor throughout training. GenAC improves value approximation, ranking reliability, and out-of-distribution generalization, and these gains translate into stronger downstream RL performance than both value-based and value-free baselines. Overall, our results suggest that stronger value modeling is a promising direction for improving credit assignment in LLM reinforcement learning.
\end{abstract}

%% file: sections/introduction.tex
Reinforcement learning (RL) has emerged as a cornerstone of large language model (LLM) post-training, transforming next-token predictors into capable assistants that solve complex problems~\mbox{\citep{jaech2024openai,guo2025deepseek,team2025kimi}}. Initially popularized through human value alignment~\citep{ouyang2022training,rafailov2023direct}, RL is now widely deployed to enhance mathematical reasoning~\citep{shao2024deepseekmath,guo2025deepseek}, code generation~\citep{le2022coderl}, and agentic systems~\citep{team2025kimi}. As these applications demand longer-horizon reasoning and more complex interactions, the ability of the underlying RL algorithms to provide effective training signals becomes a critical bottleneck.

A central challenge in RL is \emph{credit assignment}: determining which decisions within a long sequence contributed to an eventual outcome~\citep{sutton1998reinforcement}. This challenge is particularly pronounced in LLM post-training, where supervision is typically limited to a single scalar reward at the end of an entire sequence. Actor-critic methods~\citep{konda1999actor} address credit assignment by learning a value function that converts sparse outcome rewards into dense, token-level advantages, providing more informative training signals for the policy. Among these methods, Proximal Policy Optimization (PPO)~\citep{schulman2017proximal} provides a practical and influential instantiation of this idea, combining stable clipped-surrogate optimization with fine-grained advantage estimation via Generalized Advantage Estimation (GAE)~\citep{schulman2015high}. This value-based recipe has been proven effective in various applications including LLM alignment~\citep{ouyang2022training}.

Yet in recent practice, PPO-style methods have fallen out of favor. Training an accurate value model at LLM scale is both difficult and computationally expensive, pushing practitioners toward simpler value-free alternatives. Methods such as Group Relative Policy Optimization (GRPO)~\citep{shao2024deepseekmath} and REINFORCE Leave-One-Out (RLOO)~\citep{ahmadian2024back} bypass the critic entirely by estimating uniform advantages from multiple sampled trajectories, and have gained wide popularity as a practical workaround that achieves competitive performance.

However, this simplicity comes at the cost of precise credit assignment. Without a value function, these methods assign uniform credit across all tokens in a trajectory, effectively reducing sequential generation to a contextual bandit problem~\citep{langford2007epoch}. This over-simplification could cap the performance ceiling of LLM post-training, and the issue grows as the field increasingly targets longer context applications such as complex reasoning, multi-turn conversations and agentic systems. This motivates us to revisit value modeling in LLM RL. Rather than viewing the recent shift toward value-free methods as evidence that critics are unnecessary, we ask: \emph{can value models be made accurate and robust enough to restore fine-grained credit assignment in practice?}

Our primary contributions are as follows.
\begin{enumerate}
    \item We analyze the limitations of standard discriminative critics in LLM RL from both theoretical and empirical perspectives. Drawing on representation complexity theory, we provide evidence that their difficulty arises in part from limited expressiveness under the one-shot scalar prediction paradigm, rather than from optimization challenges alone. Controlled scaling experiments confirm that their approximation error does not improve reliably with model size and is highly sensitive to random seeds.
    \item We propose Generative Actor-Critic (GenAC), which replaces one-shot scalar value prediction with a generative critic that performs explicit chain-of-thought reasoning before producing value estimates, directly addressing the expressiveness limitation above. We further introduce In-Context Conditioning, a mechanism that keeps the critic calibrated to the current actor throughout training.
    \item On mathematical reasoning benchmarks, GenAC achieves superior sample efficiency and continues to improve in training regimes where baselines plateau. Supporting analyses show that the generative critic ranks candidate actions more reliably, generalizes better out-of-distribution, and produces interpretable credit assignments that are consistent with reasoning quality.
\end{enumerate}

\begin{figure}[tb]
    \centering
    \includegraphics[width=\textwidth]{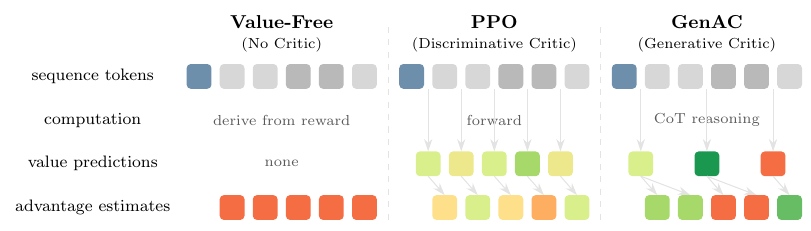}
    \caption{Comparison of advantage estimation under value-free methods, PPO, and GenAC, on a representative trajectory where initial steps are correct but later erroneous steps lead to an incorrect final answer. Value-free methods assign uniform negative advantage to all tokens, ignoring correctness of early steps; PPO enables fine-grained but noisy credit assignment due to fundamental limitations of its discriminative critic; our GenAC, equipped with a substantially stronger value estimator, assigns precise credit across segments.}
    \label{fig:intro}
\end{figure}

%% file: sections/related_works.tex
\noindent\textbf{Value-Based Algorithms.} 
Although crucial for fine-grained advantage estimation, value modeling remains underexplored beyond standard PPO in LLM RL. VC-PPO~\citep{yuan2025s} mitigates initialization bias and reward decay through value pretraining and setting $\lambda=1$ in TD($\lambda$) for value learning. We adopt these techniques, but find them insufficient for effective value learning. VRPO~\citep{zhu2025vrpo} introduces a complex information bottleneck architecture to handle noisy supervision; SAE~\citep{gong2026segmental} formulates a segment-level MDP to apply PPO on. However, all these works rely on discriminative architecture and do not isolate value function approximation as an independent factor of performance. Meanwhile, several works propose better ways of utilizing learned value models for effective policy learning~\citep{yue2025vapo,fan2025truncated,liu2025asymmetric} instead of improving value modeling. Some work~\citep{xie2025teaching,chen2025spc} involves textual critique rationales instead of value estimates; $V_0$~\citep{zhang2026v_0} learns a value model applicable only at the initial state. These works are orthogonal to ours.

\vspace{0.2em}
\noindent\textbf{Value-Free Algorithms.} 
Bandit-style algorithms such as GRPO~\citep{shao2024deepseekmath} and RLOO~\citep{ahmadian2024back} have gained popularity due to their simplicity and efficiency. While these methods avoid the cost of value modeling, they risk over-simplification and could lower the possible gains of post-training. To obtain fine-grained feedback without value models, methods like VinePPO~\citep{kazemnejad2025vineppo} and SPO~\citep{guo2025segment} use multiple continuous rollouts for unbiased segment-level value estimation. However, these sample-based estimations incur a great inference cost during training. GiGPO~\citep{feng2025group} estimates advantage by utilizing repeated states that exist only in multi-turn agent training rather than general LLM RL. Recently, OTB~\citep{li2026optimal} derives a token baseline to approximate directly.

\vspace{0.2em}
\noindent\textbf{Fine-Grained Reward Signals.} 
Another line of work tackles credit assignment by augmenting sparse outcome rewards with dense, intermediate reward signals. Process Reward Models (PRMs) provide reward at every intermediate reasoning step, requiring expensive process labels annotated by humans or rollouts~\citep{lightman2023let,wang2024math,luo2024improve}. Subsequent work proposes implicit PRMs that require only outcome labels to train~\citep{zhong2025dpo,zhong2025brite,yuan2025free,cui2025process}. However, both explicit and implicit PRMs are static evaluators highly susceptible to reward hacking and often struggle to generalize. Other approaches design heuristic reward bonuses~\citep{qu2025optimizing}. These works reformulate the RL objective with external rewards, while our work focuses on utilizing internal feedback through value estimates.

\vspace{0.2em}
\noindent\textbf{Generative Reward Models.} 
There is a growing interest in utilizing generative architectures for reward modeling. Early methods finetune language models to generate chain-of-thought reasoning before extracting probabilities from indicator tokens~\citep{zhang2024generative,mahan2024generative}, while more recent works train models with RL to directly output verbalized rewards~\citep{liu2025inference,chen2025rm,guo2025reward,sareen2025putting}. Similar ideas are also explored in PRMs~\citep{zhao2026genprm,khalifa2025process,zha2025rl}. Reward models are not relevant as they predict sparse rewards without concerning credit assignment. Comparing to process reward, value modeling is a better way to implement credit assignment: value estimates function as baseline and don't change the optimized objective in expectation, making value modeling resistant to reward hacking and robust to imperfect models. In addition, value function defines over an evolving policy, which reward functions don't. We adapt the chain-of-thought generation to the more promising and difficult value modeling task, and our approach ensures effective and grounded value learning throughout the RL process.

%% file: sections/preliminaries.tex
\noindent\textbf{Language Generation as an MDP.} 
Language generation can be modeled as a Markov Decision Process $(\mathcal{S}, \mathcal{A}, P, R, \rho, T)$, where $\mathcal{S}$ is the state space, $\mathcal{A}$ is the action space, $P$ is the transition dynamics, $R$ is the reward function, $\rho$ is the distribution of initial states, and $T$ is the maximum number of decision steps. At step $t$, the state $s_t = x \circ y_{<t} \in \mathcal{S}$ is the concatenation of the prompt $x$ and the response generated so far $y_{<t} = y_1 \circ y_2 \circ \cdots \circ y_{t-1}$; the action $a_t \in \mathcal{A}$ is the new generation at granularity of token, segment or response; the transition dynamics $P$ is deterministic, appending the new generation to the state to obtain the next state $s_{t+1} = P(s_t, a_t) = s_t \circ a_t$; the reward function $R$ assigns $0$ to all steps except the last, where it provides a binary reward depending on the correctness of the generation. A parameterized policy $\pi_\theta(a|s)$ determines the probability of taking action $a$ at state $s$. Its value function $V(s)$ is defined by the expected cumulative reward if starting at state $s$ then following $\pi$. Its Q function $Q(s, a)$ is defined by the expected cumulative reward if starting at state $s$, taking action $a$, then following $\pi$. The advantage function $A(s, a)$, defined by $Q(s, a) - V(s)$, depicts the expected improvement in cumulative reward from taking action $a$ over the average action at state $s$, then following $\pi$.

\vspace{0.2em}
\noindent\textbf{Proximal Policy Optimization.} 
PPO and its variants are the most widely used RL algorithms in LLMs. It is efficient through minibatch update and stable by optimizing a clipped surrogate objective
\begin{gather*}
    \label{eq:ppo}
    \mathcal{L}_\mathrm{PPO}(\theta) = \mathbb{E}_{x \sim \mathcal{D}, y \sim \pi_{\theta_\mathrm{old}}} \left[ -\sum_{t=1}^{|y|} \min \left( \rho_t(\theta) \hat{A}_t, \mathrm{clip} (\rho_t(\theta), 1 - \epsilon, 1 + \epsilon) \hat{A}_t \right) \right]
\end{gather*}
where $\rho_t(\theta) = \frac{\pi_\theta(a_t|s_t)}{\pi_{\theta_\mathrm{old}}(a_t|s_t)}$, $\hat{A}_t$ is the advantage estimation at step $t$, and $\epsilon$ is a small hyperparameter. Variants of PPO differ in how $\hat{A}_t$ is computed. PPO estimates token-level advantages via GAE~\citep{schulman2015high}:
\begin{gather*}
    \label{eq:gae}
    \hat{A}_t = \sum_{k=0}^{T-t} (\gamma \lambda)^k (r_{t+k} + \gamma v_\phi(s_{t+k+1}) - v_\phi(s_{t+k}))
\end{gather*}
where $v_\phi$ is a learned critic model, $r_t$ is the observed reward at step $t$, $\gamma$ is the discount factor, and $\lambda$ is a hyperparameter. The critic model is trained to minimize MSE towards TD($\lambda$) targets on the same data distribution. GRPO and RLOO estimate trajectory-level advantages utilizing the rewards of all responses for the same prompt. GRPO normalizes the rewards within a group $\hat{A}_{i,t} = \frac{r_i - \mathrm{mean}_j r_j}{\mathrm{std}_j r_j}$, while RLOO simply introduces a sample-based baseline $\hat{A}_{i,t} = r_i - \mathrm{mean}_{j, j \ne i} r_j$.

%% file: sections/limitations.tex
\label{sec:limitations:value_approximation_separation}

\begin{figure}[tb]
    \centering
    \begin{minipage}[t]{0.48\textwidth}
        \includegraphics[width=\textwidth]{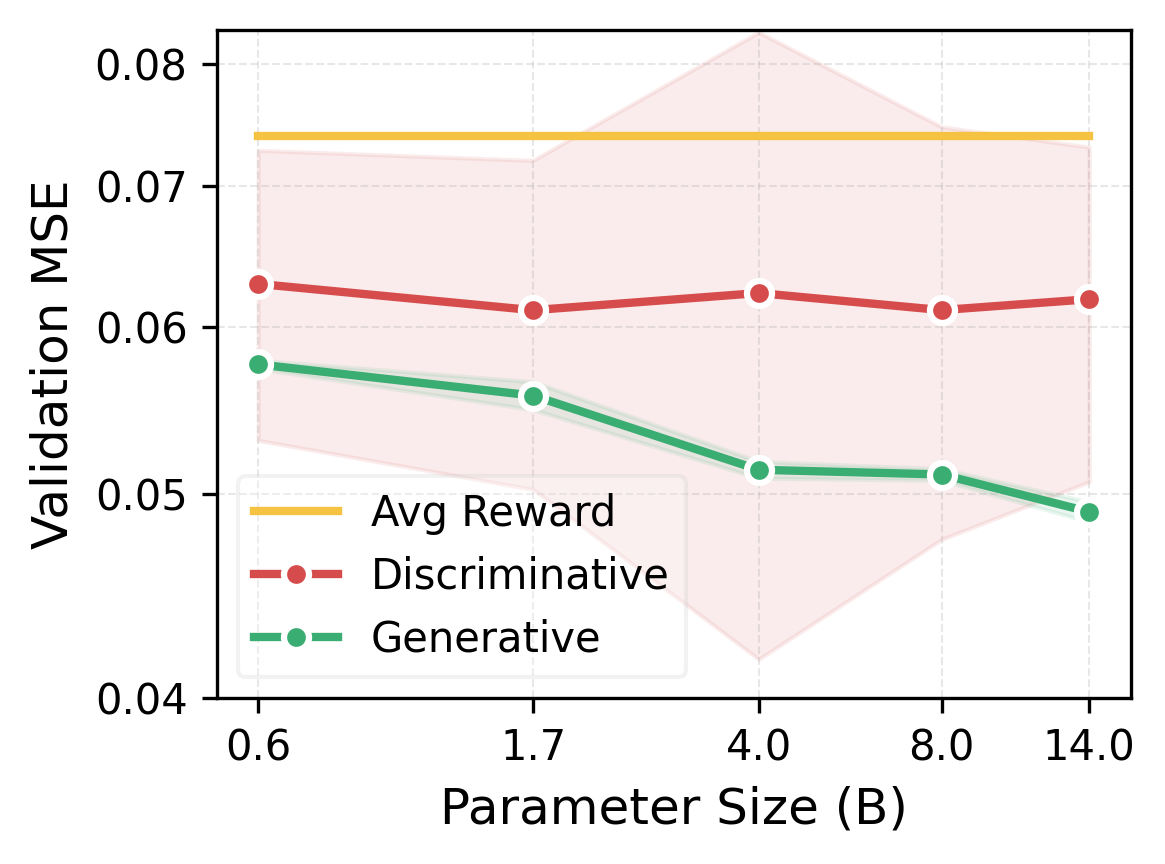}
    \end{minipage}
    \hfill
    \begin{minipage}[t]{0.48\textwidth}
        \includegraphics[width=\textwidth]{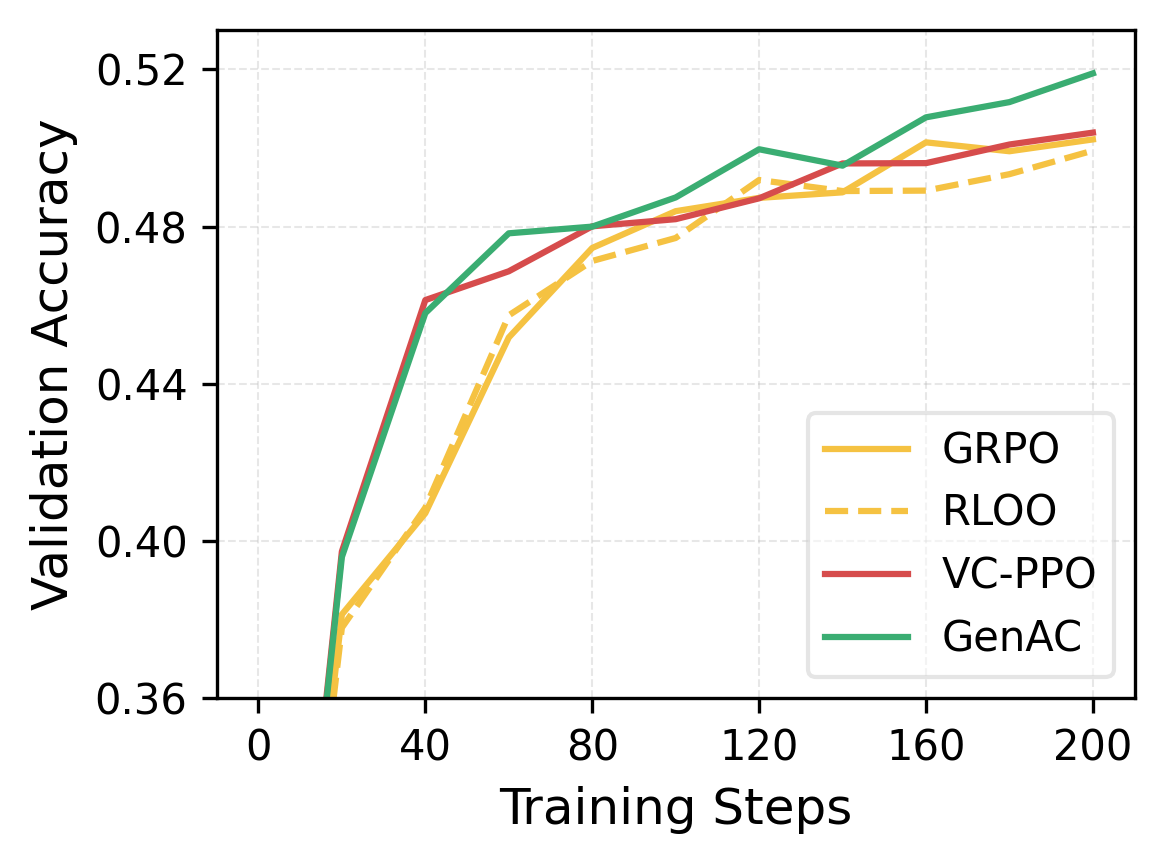}
    \end{minipage}
    \caption{
    \textbf{(a)} Comparison of approximation performance between discriminative critics and generative critics. Both critics are trained to fit the same fixed value function until convergence. We repeat each setup 10 times with different random seeds and visualize the mean and standard deviation of validation mean squared error. Notably, generative critics scale better with model size and are more robust to randomness. \texttt{Avg Reward} uses average rewards of 8 trajectories as predictions for all states, indicating insufficient approximation performance of value-free methods.
    \textbf{(b)} RL validation performance during training. Validation accuracies are \texttt{avg@16} scores averaged across six mathematical reasoning benchmarks. Value-based methods are more sample efficient than value-free methods, while generative critic outperforms discriminative critic in the long run.
    }
    \label{fig:main}
\end{figure}

We investigate the limitations of standard discriminative critics in LLM RL from both theoretical and empirical perspectives. Our analysis suggests that the difficulty of value learning is not purely an optimization problem, but is also rooted in the limited expressiveness of the one-shot scalar prediction paradigm. These findings motivate the more expressive generative critic developed in the next section.

\vspace{0.2em}
\noindent\textbf{A Representation-Complexity Perspective.} 
We study the limitations of standard discriminative critics through the lens of \emph{representation complexity}. The key issue is not only that value prediction can be noisy or difficult to optimize, but also that the value function itself may be intrinsically difficult to represent with a one-shot scalar predictor. A standard discriminative critic maps a prompt together with a partial response directly to a scalar value in a single forward pass of fixed depth. This raises a natural question: even under stable optimization, is such a predictor expressive enough for value functions arising in sparse-reward language-generation MDPs? Recent theory suggests that the answer can be no. In particular, \citep{feng2024rethinking} construct a family of MDPs parameterized by an underlying language \(L \in \mathsf{P}\), and show that the associated value-function computation can already be \(\mathsf{P}\)-complete. This is not a single pathological example, but a fairly general construction template indexed by the underlying language. Meanwhile, log-precision Transformers with constant depth and polynomial hidden dimension are confined to the complexity class \(\mathsf{TC}^{0}\)~\citep{merrill2023parallelism}, which is widely believed to be strictly weaker than \(\mathsf{P}\). The implication is that value functions can be substantially harder to represent than a standard scalar critic can express. From this perspective, the difficulty of value learning may stem not only from optimization, but also from a mismatch between the target value function and the expressive capacity of one-shot scalar prediction. Notably, since log-precision Transformers of any practical scale remain within \(\mathsf{TC}^{0}\), this perspective indicates that increasing model size alone, without adding sequential computation, should yield limited improvement, a prediction we test empirically below. With chain-of-thought, however, log-precision Transformers can characterize exactly \(\mathsf{P}\)~\citep{feng2023towards,li2024chain}, which motivates the use of chain-of-thought generation.\footnote{For readers less familiar with the complexity notation, Appendix~\ref{app:complexity_background} provides a brief introduction.}

\vspace{0.2em}
\noindent\textbf{Empirical Approximation Performance.} 
To probe this representation-complexity perspective empirically, we isolate value estimation from the rest of RL and fit a fixed value function, namely that of the initial policy \texttt{Qwen3-8B-Base}~\citep{yang2025qwen3}. Analogous to the value pretraining stage in \citep{yuan2025s}, we freeze the actor and train the critic on a held-out split of the DeepScaleR~\citep{deepscaler2025} prompt dataset. All critics are initialized directly from the Qwen3 base series. Following \citep{yuan2025s}, we formulate the MDP at the token level and fit Monte Carlo returns. We evaluate on a curated benchmark of state-value pairs, where each state consists of a validation prompt together with a partial response generated by the initial policy and truncated at semantic boundaries. Following \citep{wang2024math}, we estimate the ground-truth value of each state by averaging the final rewards over multiple continuations, and report the mean squared error between these labels and the critic predictions. To study scaling behavior, we train critics across model sizes ranging from $0.6$B to $14$B. For each size, we tune the learning rate and the amount of training data to obtain stable convergence. We repeat each setup 10 times with different random seeds and report the mean and standard deviation of the final validation MSE in Figure~\ref{fig:main}(a).

\begin{itemize}
    \item \textbf{Not scalable.} Despite substantial increases in pretrained model capacity and training compute, critics across all evaluated sizes achieve remarkably similar approximation errors. This suggests that larger discriminative critics still fail to realize materially better value representations.
    \item \textbf{Not robust.} Final performance varies sharply across random seeds for every model size. Since LLMs are typically much less sensitive to such randomness, this variation points to pronounced instability in discriminative value approximation.
\end{itemize}


Together, these results suggest that reliable value estimation may require more sequential computation than discriminative critics can provide. This limitation stems from a fundamental expressiveness barrier that scaling model size alone cannot overcome, motivating the intermediate-reasoning critic architecture developed in the next section.

%% file: sections/method.tex
\begin{figure}[tb]
    \centering
    \includegraphics[width=\textwidth]{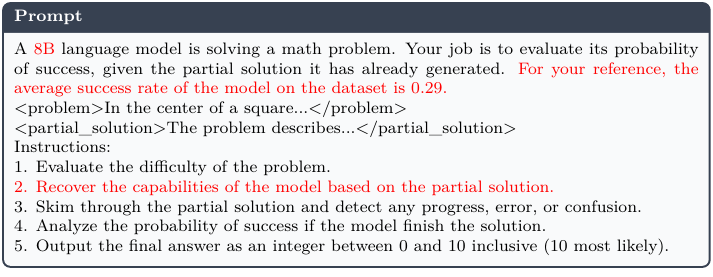}
    \caption{Our proposed prompt template. ICC hints are colored in red.}
    \label{fig:template}
\end{figure}

The representation-complexity perspective developed in Section~\ref{sec:limitations} clarifies why value modeling may require a more expressive critic parameterization. The appeal of a generative critic is not simply that it produces text before a scalar value, but that it preserves autoregressive generation and thereby enables intermediate reasoning. Recent work further suggests that such intermediate generation can strictly expand transformer expressive power relative to one-shot prediction~\citep{feng2023towards,li2024chain}. If value modeling is fundamentally constrained by representation complexity, then this additional reasoning channel provides exactly the kind of computation that a critic should exploit.

\subsection{Generative Critic}

We now describe two core components of our generative critic: an autoregressive architecture that enables chain-of-thought reasoning before value estimation, and an in-context mechanism for explicit policy conditioning. Together, they form the foundation for the training pipeline in Section~\ref{sec:method:training}.

\noindent\textbf{Architecture.} 
Standard discriminative critics replace the pretrained language modeling head with a linear projection layer that maps the final hidden states directly to a scalar value prediction. In contrast, our proposed generative critic retains the original language modeling head, fully leveraging pretrained knowledge and allowing the critic to perform explicit chain-of-thought reasoning before arriving at a final value estimate. Instead of prompting the model to verbalize an arbitrary real-valued scalar, we ask the critic to produce an integer score from $0$ to $10$ representing the likelihood of success, which we parse and normalize into a value prediction between $0$ and $1$. This scoring format is inherently more natural for a language model to reason over and predict. We formulate the MDP at the segment level: each response is split into contiguous segments based on heuristic rules, and each segment is treated as a single action. We illustrate the advantage estimation computation of value-free methods, discriminative critics, and generative critics in Figure~\ref{fig:intro}.

\vspace{0.2em}
\noindent\textbf{In-Context Conditioning (ICC).} 
A fundamental distinction between reward and value functions is that the latter are conditioned on policies, suggesting that critics must be \textit{aware} of the specific capabilities of the actor. While discriminative critics encode all relevant information in their weights, generative critics, as LLMs with in-context learning capabilities, can additionally exploit explicit policy descriptors provided in context. To this end, we design a prompt template that a) explicitly prompts the critic to infer the actor's capabilities from the partial response, and b) provides additional information about the actor, including its model size and a smoothed running average of achieved rewards, as shown in Figure~\ref{fig:template}. This grounds the critic's reasoning, making a general function approximator more tightly policy-conditioned without requiring it to encode all related information in weights alone. This is particularly important because, as we show in Section~\ref{sec:exp:ablation}, even a strong LLM like GPT-5 fails to serve as an accurate value model without sufficient knowledge of the actor. Our ablation confirms that ICC yields consistent improvements across all training stages.

\subsection{Training Pipeline}
\label{sec:method:training}

Prior to the main RL loop, the generative critic undergoes a two-stage pretraining process: (1) \textit{Supervised Finetuning} (SFT), which instills foundational reasoning patterns using synthesized reasoning traces, and (2) \textit{RL Pretraining}, which grounds the critic's reasoning in empirical, ground-truth returns. Following these stages, the critic is fully initialized and ready for joint training with the actor.

\vspace{0.2em}
\noindent\textbf{SFT.} 
To instill the reasoning capabilities into the generative critic, we construct a dataset of high-quality reasoning traces for cold-start SFT. Using a held-out split of the prompt dataset, we sample partial responses generated by the initial actor, apply the critic prompt template in Figure~\ref{fig:template}, and query GPT-5~\citep{singh2025openai} to generate reasoning traces. We collect these traces and perform standard SFT on the base model. Notably, despite being a strong language model, GPT-5 is a weak value approximator without sufficient knowledge of the evaluated policy, as Table~\ref{tab:ablation} shows. Thus the purpose of this stage is solely to equip the critic with the reasoning patterns needed for value prediction, while keeping its chain-of-thought within a controlled token budget, rather than distilling value modeling skills from the oracle, making this stage beneficial but not necessary.

\vspace{0.2em}
\noindent\textbf{RL Pretraining.} 
While SFT successfully instills reasoning capabilities, the resulting critic struggles to produce accurate value estimates due to two reasons. First, SFT models tend to overfit to stylistic patterns in the data and generalize poorly to unseen states~\citep{tajwar2024preference,chu2025sft}. Second, the critic's performance is inherently upper-bounded by the oracle, which cannot perfectly simulate the initial policy's true performance due to insufficient in-context information (details in Section~\ref{sec:exp:ablation}). To bridge this gap, we freeze the actor and train the critic via RL, directly grounding its reasoning in empirical returns. Specifically, we train the critic using REINFORCE~\cite{williams1992simple} to optimize a rule-based reward. For each state $s$, the critic generates a reasoning trace $z$ yielding a parsed value prediction $\hat{v}$, which is evaluated against the observed reward $r$: $R_v(s, z) = 1 - (r - \hat{v})^2$. If parsing $\hat{v}$ from $z$ fails, we set the reward to $0$. This reward is motivated by the MSE minimization objective used to train discriminative critics in standard PPO, while additionally enforcing valid output format. Following \citep{yuan2025s}, we set GAE $\lambda=1$ and remove the KL penalty. Training stops when validation MSE saturates.

\vspace{0.2em}
\noindent\textbf{RL Joint Training.} 
Once pretrained, the generative critic is integrated into the downstream RL loop. The procedure is identical to RL pretraining, except that the actor is now updated via PPO using advantages computed from the critic's predictions. By alternating between actor and critic updates, the critic continuously adapts its reasoning to track the shifting value function of the evolving actor, ensuring credit assignment remains accurate as the training proceeds. A detailed pseudocode is provided in Appendix~\ref{app:implementation}.

\vspace{0.2em}
\noindent\textbf{Cost Analysis.} 
We conduct a detailed cost analysis in Appendix~\ref{app:cost}. Compared to standard PPO, GenAC incurs approximately $2.1\times$ the computational cost per iteration. However, as we show in Section~\ref{sec:exp:rl}, the overhead is offset by improved sample efficiency and continued performance gains when a discriminative critic saturates. Additionally, this cost is an implementation choice, as we choose to explore what is achievable with the strongest value model we can build. We discuss how to reduce cost within our framework, and  leave further cost-optimal tuning to future work.

%% file: sections/experiments.tex
\subsection{Approximation Performance}

We repeat the controlled approximation experiment from Section~\ref{sec:limitations:value_approximation_separation}, now training generative critics under the same protocol. As shown in Figure~\ref{fig:main}(a), generative critics achieve substantially lower approximation error than their discriminative counterparts at every model size. Moreover, their error decreases consistently as model size increases and exhibits markedly lower variance across random seeds, addressing both failure modes identified in Section~\ref{sec:limitations}.

\subsection{RL Experiment}
\label{sec:exp:rl}

We now present our primary evaluation, conducting extensive RL training experiments on mathematical reasoning to assess the effectiveness of our method.

\vspace{0.2em}
\noindent\textbf{Setup.} 
We adopt a zero RL setup~\citep{guo2025deepseek}, initializing the policy with \texttt{Qwen3-8B-Base}~\citep{yang2025qwen3} and the critic with our pretrained generative critic as detailed in Section~\ref{sec:method}. Models are trained on the DeepScaleR dataset~\citep{deepscaler2025} and evaluated across six standard mathematical reasoning benchmarks: AIME 2024, AMC 2023, MATH 500~\citep{hendrycksmath2021}, AIME 2025, OlympiadBench~\citep{he2024olympiadbench}, and Minerva Math~\citep{lewkowycz2022solving}. For all evaluations, we report the average accuracy over 16 sampled generations per problem.

\vspace{0.2em}
\noindent\textbf{Baselines.} 
We compare GenAC against three RL baselines: GRPO, RLOO, and VC-PPO. As PPO variants, all algorithms share the same underlying policy optimization framework and differ solely in their advantage estimation mechanisms. GRPO and RLOO serve as value-free baselines that estimate advantages using sample scores, while VC-PPO leverages a pretrained discriminative critic.

\vspace{0.2em}
\noindent\textbf{Performance of GenAC.} 
From Figure~\ref{fig:main}(b) and Table~\ref{tab:rl}, our proposed GenAC achieves the most favorable final performance and training dynamics. Notably, it is \textit{sample-efficient}, requiring fewer training steps to outperform baselines. Furthermore, its \textit{performance margin over all baselines continues to grow} as training progresses into later half. We attribute this success to the generative critic that provides fine-grained feedback compared to bandit-style baselines, and enables more accurate credit assignment than the standard discriminative critic in VC-PPO.

\vspace{0.2em}
\noindent\textbf{The Role of Value Modeling.} 
Our comparison of advantage estimators reveals a clear hierarchy in policy performance: GenAC outperforms VC-PPO, which in turn outperforms the bandit-style baselines. Benefiting from fine-grained feedback, both GenAC and VC-PPO are more sample-efficient than GRPO or RLOO during the early stages of training. However, constrained by the limitations of its discriminative critic, VC-PPO's advantage diminishes over time, ultimately resulting in a policy only marginally above the value-free baselines. In contrast, leveraging a more expressive generative critic, GenAC sustains this momentum and achieves an increasing performance margin. This is also supported by our multi-seed replications in Appendix~\ref{app:robustness}: GRPO and VC-PPO perform similarly across multiple runs, while GenAC is clearly distinguished from all. These observations suggest that improvements in value modeling \textit{can} translate to superior policy performance in LLM RL, demonstrating that explicit, accurate value modeling remains a critical driver for effective and efficient policy optimization.

\begin{table}[tb]
    \caption{Final performance in \texttt{avg@16} accuracy (\%) on validation benchmarks.}
    \label{tab:rl}
    \centering
    \begin{tabular}{c cccccc c}
        \toprule
        Algorithm & AIME24 & AMC23 & MATH500 & AIME25 & Olympiad & Minerva & Average \\ \midrule
        GRPO & 26.46 & 69.69 & 87.20 & \textbf{22.92} & 52.92 & 42.12 & 50.22 \\
        RLOO & 24.79 & 68.75 & \textbf{87.64} & 21.25 & 53.94 & 43.22 & 49.93 \\
        VC-PPO & 25.83 & 70.78 & 87.54 & 21.67 & 53.50 & 43.02 & 50.39 \\
        GenAC & \textbf{30.00} & \textbf{74.53} & 87.48 & 21.88 & \textbf{54.16} & \textbf{43.34} & \textbf{51.90} \\
        \bottomrule
    \end{tabular}
\end{table}

\subsection{Analysis}

Having established the RL training performance, we now take a closer look at the generative critic itself, probing its design choices and characteristic properties through a series of targeted analyses.

\subsubsection{Ablation Studies}
\label{sec:exp:ablation}

To disentangle the contributions of the training stages and the ICC mechanism, we conduct an ablation study evaluating value function approximation performance and summarize the results in Table~\ref{tab:ablation}.

\vspace{0.2em}
\noindent\textbf{The Challenge of Zero-Shot Value Estimation.} 
A critical observation is that even GPT-5, the oracle model we use to synthesize reasoning traces, struggles to provide accurate value estimates by prompting alone, standing in sharp contrast to the success of LLM-as-a-Judge in reward modeling. We attribute this discrepancy to the fundamental requirement of policy-conditioning. Without explicit training, even a strong generalist model fails to accurately predict the value of a specific policy, highlighting the necessity of our specialized training pipeline.

\vspace{0.2em}
\noindent\textbf{Impact of Training Stages.} 
While SFT provides a large improvement over the base model by instilling the reasoning patterns needed for value prediction within a controlled token budget, its predictions are bounded by the limited accuracy of the oracle itself. RL pretraining is therefore necessary, as it enables continuous improvement by optimizing directly toward observed returns. We further note that the main benefit of the SFT stage is yielding a stable reasoning length rather than improving prediction accuracy, which RL pretraining alone is sufficient to improve (Appendix~\ref{app:sft_ablation}). Thus the SFT stage is beneficial but not strictly necessary.

\vspace{0.2em}
\noindent\textbf{Efficacy of ICC.} 
We observe that removing ICC consistently degrades performance across all configurations. Although ICC alone is insufficient to transform a generalist LLM into an accurate value model, this consistent benefit indicates that it provides additional information that the model weights alone cannot fully capture, effectively enhancing the precision of value estimation.

\begin{table}[tb]
    \begin{minipage}[t]{0.48\linewidth}
        \centering
        \caption{Ablation on training pipeline and ICC in approximation performance (MSE).}
        \label{tab:ablation}
        \begin{tabular}{lrr}
            \toprule
            Model & w/ ICC & w/o ICC \\ \midrule
            GPT-5 & 0.129 & 0.164 \\ \midrule
            Qwen3-8B-Base & 0.299 & 0.308 \\
            + \textit{SFT} & 0.137 & 0.164 \\
            + \textit{RL} & 0.051 & 0.058 \\
            \bottomrule
        \end{tabular}
    \end{minipage}
    \hfill
    \begin{minipage}[t]{0.48\linewidth}
        \centering
        \caption{Relative ranking performance of different critics in top-1 accuracy.}
        \label{tab:rr}
        \begin{tabular}{cccc}
            \toprule
            \# Candidates & Discriminative & Generative \\ \midrule
            2 & 0.604 & 0.711 \\
            4 & 0.388 & 0.492 \\
            8 & 0.204 & 0.332 \\
            \bottomrule
        \end{tabular}
    \end{minipage}
\end{table}

\subsubsection{Relative Ranking Performance}

While approximation error measures the absolute performance of value models, RL optimization depends heavily on relative comparisons due to mechanisms like advantage estimation and group sampling. Therefore, assessing value prediction quality requires a relative ranking probe.

\vspace{0.2em}
\noindent\textbf{Setup.} 
To evaluate this capability, we generate multiple candidate segments per prompt, compute their ground-truth values using the average score of Monte Carlo rollouts, and score them with either discriminative or generative critic. We measure the top-1 accuracy, i.e. the ratio of prompts where the critic's highest-scoring candidate matches the candidate with the highest ground-truth value, and report the results in Table~\ref{tab:rr}.

\vspace{0.2em}
\noindent\textbf{Results.} 
The results demonstrate that the generative critic consistently outperforms its discriminative counterpart across all candidate pool sizes. Crucially, the performance gap widens as the decision space grows larger: the discriminative critic's accuracy degrades quickly towards random chance while the generative critic maintains a much higher accuracy. This indicates that in addition to yielding better absolute approximation, the generative critic provides a more reliable advantage signal for distinguishing between possible actions.

\subsubsection{Generalization}

\begin{table}[tb]
    \caption{Generalization performance of generative critics over discriminative critics, measured by the approximation MSE reduction on datasets of varying distribution shift.}
    \label{tab:ood}
    \centering
    \begin{tabular}{cccccr}
        \toprule
        Dataset & Distribution Shift & Discriminative & Generative & Reduction \\ \midrule
        DeepScaleR & None & 0.0666 & 0.0531 & 20.3\% \\
        Math500 & Mild & 0.1311 & 0.0796 & 39.3\% \\
        AIME24 & Moderate & 0.0524 & 0.0249 & 52.5\% \\
        GPQA & High & 0.0583 & 0.0300 & 48.5\% \\
        \bottomrule
    \end{tabular}
\end{table}

\begin{figure}[tb]
    \centering
    \includegraphics[width=\textwidth]{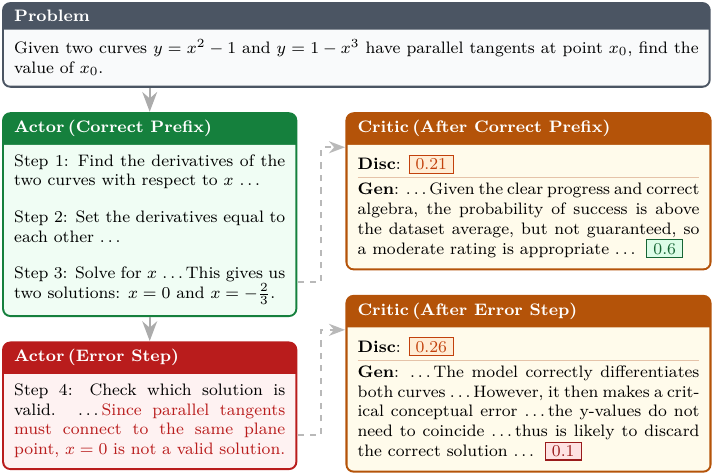}
    \caption{A case where the generative critic's accurate value estimates translate into detecting a conceptual error, while the discriminative critic provides less informative and misleading assessments.}
    \label{fig:case}
\end{figure}

Although value models are mostly responsible for providing accurate feedback on in-distribution training data, their out-of-distribution generalization is crucial for effective and stable learning on exploratory off-policy data. A critic that overfits superficial in-distribution patterns will provide noisy or degenerate feedback that hinders the learning process and could lead to premature convergence.

\vspace{0.2em}
\noindent\textbf{Setup.} 
To assess this robustness, we apply our approximation performance evaluation to the pretrained critics across a spectrum of datasets with increasing degrees of distributional shift. Concretely, we progressively move from exact in-distribution training (DeepScaleR) to mathematical problems of varying difficulty (Math500 for easier and AIME24 for harder), and finally to a distinct domain of general scientific reasoning (GPQA~\citep{rein2024gpqa}).

\vspace{0.2em}
\noindent\textbf{Results.} 
The evaluation in Table~\ref{tab:ood} reveals that the two architectures differ markedly in generalization. While the generative critic achieves a modest MSE reduction on in-distribution data compared to the discriminative baseline, its relative advantage increases significantly as the distributional shift increases, reaching over 50\% reduction on AIME24. This suggests that the generative critic evaluates reasoning quality based on transferable patterns rather than distribution-specific memorization. This stronger generalization helps explain the growing performance gap between GenAC and VC-PPO in Figure~\ref{fig:main}(b): as the actor evolves away from its initialization, the generative critic continues to provide reliable feedback while the discriminative critic degrades.

\subsubsection{Case Study}

We illustrate a qualitative case study in Figure~\ref{fig:case}. The problem requires finding the $x$-coordinate at which two curves possess parallel tangents. Following a correct prefix, the generative critic acknowledges this progress by assigning a value of $0.6$. In Step 4, however, the actor commits a conceptual error by dismissing a valid solution. The generative critic accurately points out this misconception and assigns a low value of $0.1$. Notably, since these predictions are independent, the abrupt value drop arises from two precise assessments. In contrast, the discriminative critic yields less informative and misleading predictions. This example demonstrates the generative critic's ability to assign credit both accurately and interpretably.

%% file: sections/conclusion.tex
This work re-establishes learned value models as a critical driver of effective LLM post-training by identifying a key source of their representational instability as a limitation of discriminative architectures, addressing it through a generative critic equipped with chain-of-thought reasoning and In-Context Conditioning, and validating with extensive experiments in mathematical reasoning benchmarks. Its limitations concern cost and the theory-practice gap. The paper focuses on demonstrating what is possible with accurate credit assignment, pointing out techniques to reduce cost but leaving optimal cost tuning as future work. The complexity motivation requires polynomial reasoning length that is practically bounded. Future directions include extension beyond mathematical reasoning, decomposing and designing better In-Context Conditioning prompt, and deciding optimal segmentation granularity. The importance of credit assignment grows as LLM post-training continues in longer horizon and more complex scenarios, and we believe expressive value modeling is a natural and promising foundation for effective and efficient optimization.

%% file: appendices/theory.tex
\noindent\textbf{Complexity Classes \(\mathsf{TC}^{0}\) and \(\mathsf{P}\).} 
The complexity class \(\mathsf{P}\) denotes problems solvable in polynomial time, which may still require many sequential computational steps. In contrast, \(\mathsf{TC}^{0}\) denotes computations realizable by polynomial-size, constant-depth threshold circuits, which are far more parallel and far less sequential. In particular, prior work shows that log-precision Transformers with constant depth and polynomial hidden dimension are confined to the complexity class \(\mathsf{TC}^{0}\)~\citep{merrill2023parallelism}. The widely held assumption \(\mathsf{TC}^{0} \neq \mathsf{P}\) therefore says that some polynomial-time computations cannot be compressed into a constant number of layers.

\vspace{0.2em}
\noindent\textbf{Value-Focused Restatement from~\citep{feng2024rethinking}.} 
A value-focused restatement of their result is as follows. They construct a family of \(\mathsf{P}\) MDPs indexed by an underlying language \(L \in \mathsf{P}\), so the result applies to a broad class of constructions rather than a single example. When \(L\) is \(\mathsf{P}\)-complete, computing the optimal value function for the corresponding MDP family is itself \(\mathsf{P}\)-complete. In other words, even for an explicitly specified finite MDP, evaluating the correct value at a state can already capture the full difficulty of arbitrary polynomial-time computation. Their expressiveness result makes the implication for standard neural predictors more concrete: under the assumption \(\mathsf{TC}^{0} \neq \mathsf{P}\), these value functions cannot be represented by constant-depth, polynomial-width MLPs or Transformers. The takeaway is that value computation may intrinsically require more sequential reasoning than a shallow one-shot predictor can express.

%% file: appendices/implementation.tex
\begin{algorithm}[tb]
    \caption{GenAC}
    \label{alg:main}
    \begin{algorithmic}[1]
    \Require Actor $\pi_\theta$, generative critic $V_\phi$, prompt dataset $\mathcal{D}$, actor verifier $R$, critic verifier $R_v$, ICC update momentum $c$, batch size $B$, minibatch size $b$
    \State Initialize ICC hint $\bar{r} \leftarrow 0$
    \For{each training iteration}
        \State Sample a batch of prompts $\{x^i\}_{i=1}^B \sim \mathcal{D}$
        \For{each prompt $x^i$}
            \State Sample response $y^i \sim \pi_\theta (\cdot | x^i)$
            \State Compute reward $r^i \leftarrow R(x^i, y^i)$
            \State Update ICC hint $\bar{r} \leftarrow c \cdot \bar{r} + (1 - c) \cdot r^i$
            \State Compute old probabilities $\pi_\mathrm{old}(y^i | x^i) \leftarrow \pi_\theta(y^i | x^i)$
        \EndFor
        \For{each minibatch $\{(x^j, y^j, r^j, \pi_\mathrm{old}(y^j | x^j))\}_{j \in [b]}$}
            \For{$j=1,2,\ldots,b$}
                \State Split response into segments $\{y^j_k\}_{k=1}^{T_j} \leftarrow \mathrm{Segment}(y^j)$ \label{line:segment}
                \For{$k=1,2,\ldots,T_j$}
                    \State Apply ICC prompt template $s^j_k \leftarrow \mathrm{Template}(x^j, y^j_{:k}, \bar{r})$
                    \State Sample response from critic $z^j_k \sim V_\phi(\cdot | s^j_k)$
                    \State Parse value prediction $\hat{v}^j_k \leftarrow \mathrm{Parse}(z^j_k)$
                    \State Compute critic reward $r^j_k \leftarrow R_v(z^j_k, r^j)$
                \EndFor
                \State Broadcast to token value $\{\hat{v}_t^j\}_{t=1}^{|y^j|} \leftarrow \mathrm{Broadcast}([\hat{v}^j_1, \hat{v}^j_2, \ldots, \hat{v}^j_{T_j}], [|y_1^j|, |y_2^j|, \ldots, |y_{T_j}^j|])$ \label{line:broadcast}
                \State Compute actor advantage $\{\hat{A}_t^j\}_{t=1}^{|y^j|} \leftarrow \mathrm{BatchNorm} \left( \mathrm{GAE} \left(r^j, \{\hat{v}_t^j\}_{t=1}^{|y^j|} \right) \right)$ \label{line:gae}
            \EndFor
            \State Update actor with PPO:
            \Statex \hspace{\algorithmicindent} $\begin{gathered}
                \mathcal{L}_\theta \leftarrow - \sum_{j=1}^{b} \sum_{t=1}^{|y^j|} \min \left( \frac{\pi_\theta(y_t^j | x^j \circ y_{<t}^j)}{\pi_\mathrm{old}(y_t^j | x^j \circ y_{<t}^j)} \hat{A}_t^j, \mathrm{clip} (\frac{\pi_\theta(y_t^j | x^j \circ y_{<t}^j)}{\pi_\mathrm{old}(y_t^j | x^j \circ y_{<t}^j)}, 1 - \epsilon, 1 + \epsilon) \hat{A}_t^j \right)
            \end{gathered}$
            \State Update critic with REINFORCE:
            \Statex \hspace{\algorithmicindent} $\begin{gathered}
                \mathcal{L}_\phi \leftarrow - \sum_{j=1}^b \sum_{k=1}^{T_j} \sum_{t=1}^{|z^j_k|} \log V_\phi(z_{k,t}^j | s^j_k \circ z_{k,<t}^j) r^j_k
            \end{gathered}$
        \EndFor
    \EndFor
    \end{algorithmic}
\end{algorithm}

\noindent\textbf{Pseudocode.} 
The pseudocode for GenAC joint training is presented in Algorithm~\ref{alg:main}. In line~\ref{line:segment}, the actor's response is segmented into a sequence of short segments using heuristic rules detailed below. Since group sampling is applied hence sample-based initial values are available, critics are not queried or evaluated for predictions on initial states. In line~\ref{line:broadcast}, the resulting segment-level value predictions are broadcast back to token-level to match the token-level objective in standard PPO. In line~\ref{line:gae}, following standard practice, we use GAE with $\lambda=1$ and batch normalization to compute advantage estimates.

\vspace{0.2em}
\noindent\textbf{Segmentation Rule.} 
We employ a delimiter-based heuristic to decompose a full model response into coherent segments. Specifically, we first identify candidate delimiters (periods and newline characters) and filter out any that fall within mathematical expressions. The remaining delimiters are used as segmentation boundaries. In addition, we then merge consecutive short segments until each meets a minimum length of 64 tokens, yielding an average segment length of approximately 120 tokens.

\vspace{0.2em}
\noindent\textbf{Code.} 
We develop our code based on verl~\citep{sheng2024hybridflow} with vLLM~\citep{kwon2023efficient} engine and FSDP~\citep{zhao2023pytorch} backend. All experiments are conducted on a node of Nvidia 80G A100 GPUs.

\vspace{0.2em}
\noindent\textbf{Dataset Construction.} 
All datasets source from the DeepScaleR dataset. The value benchmark in the approximation experiments includes 1024 problems, each with one response split into segments to construct 14190 states, and each state conducts 16 partial rollouts to obtain value estimate used as ground-truth target. The RL pretrain prompt dataset includes 12800 problems, or 100 RL steps. This dataset is also used as the SFT problem set. For each problem and one of its responses, we randomly select one prefix, construct the prompt with template shown in Figure~\ref{fig:template}, query GPT-5, then collect model outputs without further processing. The main RL prompt dataset includes 25600 problems, or 200 RL steps. The relative ranking experiment involves 1024 problems, each 8 candidates, whose ground-truth value is estimated through 32 partial rollouts. The generalization experiment constructs value benchmarks of different sources similarly as the DeepScaleR dataset, with an exception of using multiple responses per problem (2 for Math500, 30 for AIME24, 4 for GPQA) to roughly match the amount of total sequences.

\vspace{0.2em}
\noindent\textbf{Hyperparameters for the RL Experiment.} 
All hyperparameters are detailed in Table~\ref{tab:hyper_rl}, largely following the example scripts in verl. During validation, we sample 16 responses per problem using the same sampling parameters as during training and report their average accuracy.

\noindent\textbf{Hyperparameters for the Approximation Experiment.} 
For discriminative critics, we perform a hyperparameter search over learning rates in $\{ 1e-5, 5e-6, 4e-6, 2e-6, 1e-6 \}$ and training steps in $\{ 100, 200 \}$ per model size. For generative critics, we fix the learning rate to $1e-6$ and the number of training steps to $100$, as they are much more robust to such variation. See Table~\ref{tab:hyper_approx} for the final selected hyperparameters.

\begin{table}[tb]
    \centering
    \caption{Hyperparameters used for RL Training.}
    \label{tab:hyper_rl}
    \begin{tabular}{cc}
        \toprule
        Hyperparameter & Value \\
        \midrule
        Actor Learning Rate & 1e-6 \\
        Discriminative Critic Learning Rate & 5e-6 \\
        Generative Critic Learning Rate & 1e-6 \\
        Adam Betas & $(0.9, 0.999)$ \\
        Adam Weight Decay & 0.01 \\
        Training Steps & 200 \\
        Prompt Batch Size & 128 \\
        Group Size & 8 \\
        Micro Batch Size & 256 \\
        Maximum Prompt Length & 1024 \\
        Maximum Response Length & 8192 \\
        Maximum Critic Response Length & 1024 \\
        PPO $\epsilon$ & $0.2$ \\
        GAE $\lambda$ & $1.0$ \\
        Discount $\gamma$ & $1.0$ \\
        \bottomrule
    \end{tabular}
\end{table}

\begin{table}[tb]
    \centering
    \caption{Hyperparameters used for approximation scaling experiments.}
    \label{tab:hyper_approx}
    \begin{tabular}{ccccc}
        \toprule
        \multirow{2}{*}{Model Size} & \multicolumn{2}{c}{Discriminative Critics} & \multicolumn{2}{c}{Generative Critics} \\
        \cmidrule(lr){2-3}\cmidrule(lr){4-5}
         & Learning Rate & Training Steps & Learning Rate & Training Steps \\
        \midrule
        0.6 & 1e-5 & 100 & \multirow{5}{*}{1e-6} & \multirow{5}{*}{100} \\
        1.7 & 5e-6 & 100 &                       &                      \\
        4   & 4e-6 & 200 &                       &                      \\
        8   & 1e-6 & 200 &                       &                      \\
        14  & 1e-6 & 200 &                       &                      \\
        \bottomrule
    \end{tabular}
\end{table}

%% file: appendices/sft_ablation.tex
\begin{figure}[tb]
    \centering
    \includegraphics[width=0.6\textwidth]{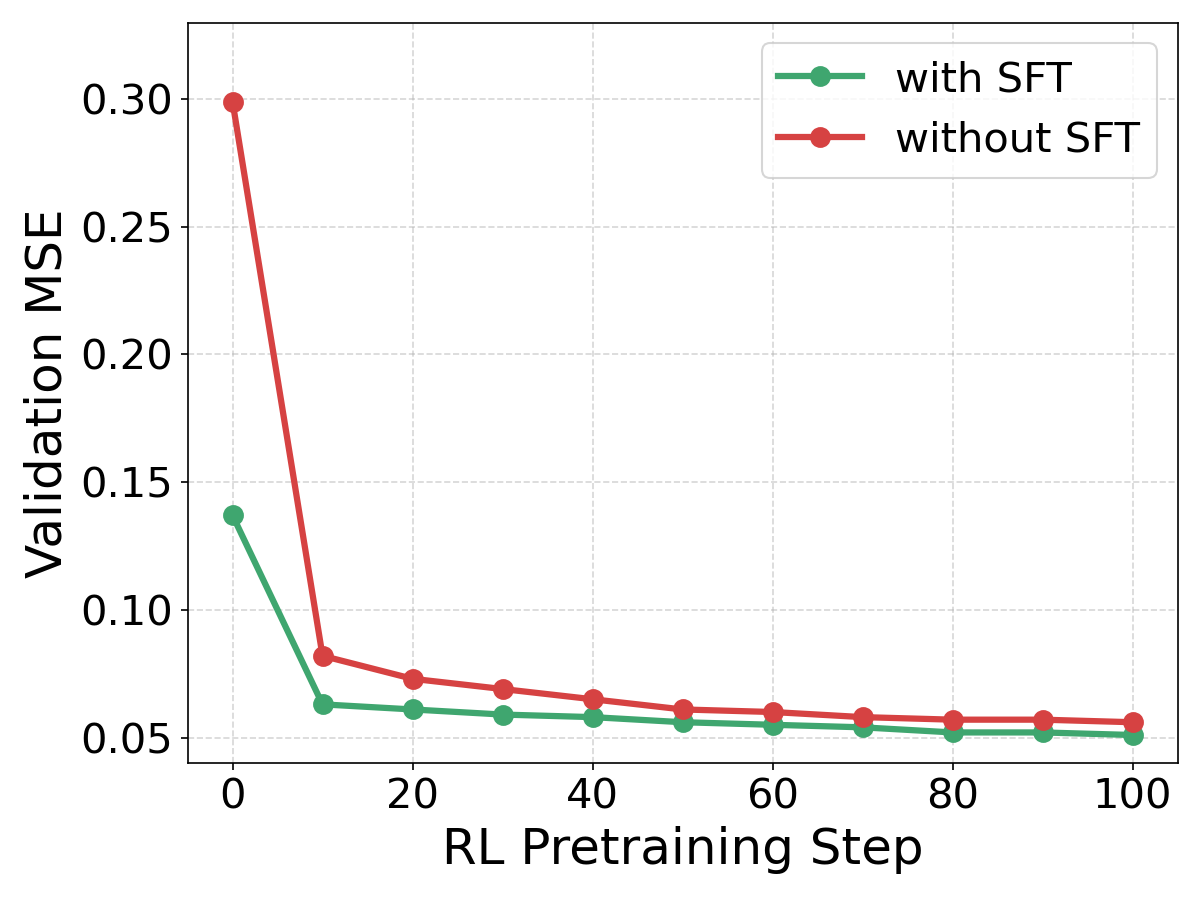}
    \caption{Validation MSE of generative critics during RL pretraining, with and without the SFT stage.}
    \label{fig:sft_ablation}
\end{figure}

\noindent\textbf{Setup.} 
To isolate the effect of the SFT stage, we skip it entirely, initializing the critic directly from the base model, and run the identical RL pretraining procedure. We report the validation MSE in Figure~\ref{fig:sft_ablation}.

\vspace{0.2em}
\noindent\textbf{Results.} 
RL pretraining alone recovers most of the MSE reduction, reaching $0.056$ at the end of training versus $0.051$ with SFT. The remaining difference lies in the stability of the critic's reasoning length: without SFT the chain-of-thought length starts longer and fluctuates substantially across training, whereas with SFT it stays within a narrow band. The SFT stage therefore serves as a natural warmup that provides a stable reasoning format and token budget, while RL pretraining is responsible for the accuracy, suggesting that in our pipeline, it is beneficial rather than necessary.

%% file: appendices/robustness.tex
\noindent\textbf{Setup.} 
We repeat the GRPO and VC-PPO with two additional random seeds, in total three runs per baseline, and report all results in Table~\ref{tab:seeds}.

\vspace{0.2em}
\noindent\textbf{Results.} 
Both GRPO and VC-PPO perform very similarly across all three runs, with average scores stable to within approximately $0.4$ points. This low run-to-run variance also indicates the robustness of both the algorithms and our evaluation protocol. GenAC is clearly distinguished from both.

\begin{table}[tb]
    \caption{Final performance in \texttt{avg@16} accuracy (\%) on validation benchmarks, including two additional runs of GRPO and VC-PPO with different random seeds.}
    \label{tab:seeds}
    \centering
    \small
    \begin{tabular}{l cccccc c}
        \toprule
        & AIME24 & AMC23 & MATH500 & AIME25 & Olympiad & Minerva & Avg \\
        \midrule
        GRPO (Table~\ref{tab:rl}) & 26.46 & 69.69 & 87.20 & 22.92 & 52.92 & 42.12 & 50.22 \\
        GRPO (seed 2) & 24.17 & 69.06 & 87.74 & 22.71 & 53.06 & 44.23 & 50.16 \\
        GRPO (seed 3) & 24.58 & 68.28 & 87.09 & 21.25 & 53.23 & 42.56 & 49.50 \\
        GRPO (mean) & 25.07 & 69.01 & 87.34 & 22.29 & 53.07 & 42.97 & 49.96 \\
        \midrule
        RLOO (Table~\ref{tab:rl}) & 24.79 & 68.75 & 87.64 & 21.25 & 53.94 & 43.22 & 49.93 \\
        \midrule
        VC-PPO (Table~\ref{tab:rl}) & 25.83 & 70.78 & 87.54 & 21.67 & 53.50 & 43.02 & 50.39 \\
        VC-PPO (seed 2) & 27.50 & 70.78 & 87.94 & 22.71 & 53.86 & 44.58 & 51.23 \\
        VC-PPO (seed 3) & 25.21 & 72.34 & 88.46 & 21.46 & 54.23 & 43.87 & 50.93 \\
        VC-PPO (mean) & 26.18 & 71.30 & 87.98 & 21.95 & 53.86 & 43.82 & 50.85 \\
        \midrule
        GenAC (Table~\ref{tab:rl}) & 30.00 & 74.53 & 87.48 & 21.88 & 54.16 & 43.34 & 51.90 \\
        \bottomrule
    \end{tabular}
\end{table}

%% file: appendices/cost.tex
We analyze the total FLOPs of GenAC in comparison to GRPO, VC-PPO and VinePPO. We begin by deriving per-step cost, and then total cost including initialization, before discussing immediate cost reduction implementations under our framework.

\vspace{0.2em}
\noindent\textbf{Per-Iteration Costs.} 
Following standard practice, we adopt the following formula to count FLOPs:
\begin{gather*}
    C_\mathrm{generation} = 2 P T \\
    C_\mathrm{forward} = 2 P T \\
    C_\mathrm{train} = 6 P T
\end{gather*}
where $P$ is the number of model parameters and $T$ is the number of processed tokens.

\vspace{0.2em}
The actor terms are common to all methods: $2P_\mathrm{A}T_\mathrm{A}$ for generation, $2P_\mathrm{A}T_\mathrm{A}$ for the old logprob forward pass and $6P_\mathrm{A}T_\mathrm{A}$ for training. GRPO uses this alone:
\begin{gather*}
    C_\mathrm{GRPO} = 2 P_\mathrm{A} T_\mathrm{A} + 2 P_\mathrm{A} T_\mathrm{A} + 6 P_\mathrm{A} T_\mathrm{A} = 10 P_\mathrm{A} T_\mathrm{A}.
\end{gather*}
PPO involves a discriminative critic of similar size to the actor, which reads the same tokens:
\begin{gather*}
    C_\mathrm{PPO} = 10 P_\mathrm{A} T_\mathrm{A} + 2 P_\mathrm{C} T_\mathrm{A} + 6 P_\mathrm{C} T_\mathrm{A} = 10 P_\mathrm{A} T_\mathrm{A} + 8 P_\mathrm{C} T_\mathrm{A}.
\end{gather*}
VC-PPO uses the same critic within the RL training and thus has the same per-iteration cost. GenAC utilizes a generative critic that generates and learns on different tokens. With average segment length $L_1$ and average critic generation length $L_2$, the critic processes the full actor trace plus  branched generations,
\begin{gather*}
    T_\mathrm{GC} = T_\mathrm{A} + \frac{T_\mathrm{A}}{L_1} L_2, \\
    C_\mathrm{GenAC} = 10 P_\mathrm{A} T_\mathrm{A} + 2 P_\mathrm{GC} T_\mathrm{GC} + 6 P_\mathrm{GC} T_\mathrm{GC} = 10 P_\mathrm{A} T_\mathrm{A} + 8 P_\mathrm{GC} T_\mathrm{GC}.
\end{gather*}
VinePPO involves branched Monte Carlo rollouts. With average branch length $L_3$ and $n$ branches per prefix,
\begin{gather*}
    T_\mathrm{vine} = \frac{T_\mathrm{A}}{L_1} L_3 n, \\
    C_\mathrm{VinePPO} = 10 P_\mathrm{A} T_\mathrm{A} + 2 P_\mathrm{A} T_\mathrm{vine}.
\end{gather*}

\vspace{0.2em}
\noindent\textbf{Total Cost.} 
The main RL training lasts $200$ steps, each $128$ problems, $8$ responses per problem, and average response length $2048$ tokens ($~1000$ at the initial stage, $~4000$ at late stage):
\begin{gather*}
    T_\mathrm{A} \approx 200 \times 128 \times 8 \times 2048 = 419.4\text{M}.
\end{gather*}
With $L_1 \approx 120$, $L_2 \approx 300$, $L_3 \approx 1024$ (half of the average response length) and $n = 4$:
\begin{gather*}
    T_\mathrm{GC} = T_\mathrm{A} + \frac{T_\mathrm{A}}{L_1} L_2 \approx 1467.9\text{M}, \\
    T_\mathrm{vine} = \frac{T_\mathrm{A}}{L_1} L_3 n \approx 14315.5\text{M}.
\end{gather*}
The discriminative value pretraining uses $200$ steps and average initial response length of $1000$ tokens:
\begin{gather*}
    T_\mathrm{VPD} \approx 128 \times 8 \times 200 \times 1000 = 204.8\text{M}.
\end{gather*}
SFT uses total trajectory count $12800$, average response length $1000$ tokens, and average critic length $300$ tokens:
\begin{gather*}
    T_\mathrm{SFT} \approx 12800 \times \left( \frac{1000}{2} + 300 \right) = 10.2\text{M}.
\end{gather*}
The generative value pretraining uses $100$ steps, average initial response length $1000$ tokens, average segment length of $120$ tokens, and an average critic length of $300$ tokens:
\begin{gather*}
    T_\mathrm{VPG} \approx 128 \times 8 \times 100 \times \left( 1000 + \frac{1000}{120} \times 300 \right) = 358.4\text{M}.
\end{gather*}

\vspace{0.2em}
Plugging in and omitting $P_\mathrm{A} \approx P_\mathrm{C} \approx P_\mathrm{GC}$, the total cost of each method, including initialization, is
\begin{gather*}
    C_\mathrm{GRPO} = 10 T_\mathrm{A} \approx 4194\text{M}, \\
    C_\mathrm{PPO} \approx 18 T_\mathrm{A} \approx 7549.2\text{M}, \\
    C_\mathrm{VC\text{-}PPO} \approx 6 T_\mathrm{VPD} + 18 T_\mathrm{A} \approx 1228.8\text{M} + 7549.2\text{M} = 8778\text{M}, \\
    C_\mathrm{GenAC} \approx 6 T_\mathrm{SFT} + 8 T_\mathrm{VPG} + 10 T_\mathrm{A} + 8 T_\mathrm{GC} \approx 61.2\text{M} + 2867.2\text{M} + 4194\text{M} + 11743.2\text{M} = 18866\text{M}, \\
    C_\mathrm{VinePPO} = 10 T_\mathrm{A} + 2 T_\mathrm{vine} \approx 4194\text{M} + 28631\text{M} = 32825\text{M}.
\end{gather*}
Initialization accounts for  $14\%$ of VC-PPO's total cost and $16\%$ of GenAC's. Both the per-step and total cost are summarized in Table~\ref{tab:cost}. We note that we choose not to tune GenAC's cost, which can be heavily reduced within our framework as discussed below.

\begin{table}[tb]
    \centering
    \caption{Summary of per-iteration relative cost and total cost including one-time initialization. GenAC's cost is intentionally untuned, and it can be reduced substantially under our framework.}
    \label{tab:cost}
    \small
    \begin{tabular}{lcc}
        \toprule
        Method & Per iteration & Total FLOPs (omit $P$) \\
        \midrule
        GRPO & $0.6 \times$ & $4194$M \\
        PPO & $1.0 \times$ & $7549.2$M \\
        VC-PPO & $1.0 \times$ & $8778$M \\
        GenAC & $2.1 \times$ & $18866$M \\
        VinePPO & $4.3 \times$ & $32825$M \\
        \bottomrule
    \end{tabular}
\end{table}

\vspace{0.2em}
\noindent\textbf{Reducing the Cost.} 
The one-time initialization cost can be reduced as the SFT stage can be skipped and the value pretraining MSE curve saturates early (Appendix~\ref{app:sft_ablation}), so both can be shortened with little performance loss. The dominant term in RL training is the generative critic inference and training cost ($8PT_\mathrm{GC} \approx 2.8\times$ GRPO) that depends on two hyperparameters: the segmentation strategy, affecting the average segment length ($L_1$), and the critic decode budget, affecting the average critic response length ($L_2$). Our deliberate choice of near-maximal feedback density (effectively about $33$ segments per $4000$-token response at late RL training) makes $T_\mathrm{GC}$ high. With coarser segmentation, the cost can be significantly reduced (Table~\ref{tab:cost_granularity}).

In addition, generation at fine granularity is required for dense actor feedback, but training the critic on every generated sequence is a choice rather than a requirement. Our implementation fixes the number of critic minibatches to keep off-policyness constant, absorbing the varying sequence count into a minibatch size that easily reaches $256 \times 33$, over an order of magnitude above standard choices of $256$. Therefore, subsampling the critic's training set while generating at full granularity preserves the actor feedback density while drastically cutting the cost (Table~\ref{tab:cost_batch}).

Both tables are arithmetic under our cost model, and the cost is adjustable along both axes, decreasing asymptotically to PPO.

\begin{table}[tb]
    \centering
    \caption{GenAC per-iteration cost under coarser feedback granularity.}
    \label{tab:cost_granularity}
    \small
    \begin{tabular}{ccc}
        \toprule
        \# Feedback & $T_\mathrm{GC} / T_\mathrm{A}$ & Relative cost \\
        \midrule
        33 & 3.5 & $3.8 \times$ \\
        10 & 1.75 & $2.4 \times$ \\
        5 & 1.375 & $2.1 \times$ \\
        1 & 1.075 & $1.9 \times$ \\
        \bottomrule
    \end{tabular}
\end{table}

\begin{table}[tb]
    \centering
    \caption{GenAC per-iteration cost with subsampled critic training set..}
    \label{tab:cost_batch}
    \small
    \begin{tabular}{ccc}
        \toprule
        \# Feedback & \# Critic batch size ($\times 256$) & Relative cost \\
        \midrule
        33 & 10 & $2.8 \times$ \\
        33 & 5 & $2.5 \times$ \\
        33 & 1 & $2.3 \times$ \\
        10 & 5 & $2.2 \times$ \\
        10 & 1 & $2.0 \times$ \\
        5 & 1 & $1.9 \times$ \\
        \bottomrule
    \end{tabular}
\end{table}

\vspace{0.2em}
\noindent\textbf{Summary.} 
Table~\ref{tab:cost} summarizes the per-iteration relative cost and the total cost including one-time initializations. Value-based methods are generally costlier than value-free alternatives, but this is compensated by increased sample efficiency and continued gains in the long run (Section~\ref{sec:exp:rl}). Additionally, the cost of GenAC is an implementation choice, as we choose to explore what is achievable with the strongest value model we can build. We discuss how to reduce cost within our framework, and  leave further cost-optimal tuning to future work.

%% file: appendices/broader_impact.tex
This work proposes a reinforcement learning algorithm for training large language 
models. Potential positive impacts include improved reasoning capabilities and 
alignment of LLMs, which could benefit a wide range of downstream applications. 
However, more effective LLM training methods could also accelerate the development 
of more capable models, which carries dual-use risks if applied to harmful objectives. 
We note that this work is foundational research, and we encourage responsible 
deployment practices in any downstream application.